\documentclass{article}

\usepackage[final]{neurips_2025}
\usepackage{graphicx}
\usepackage{algorithm}
\usepackage[noend]{algpseudocode}
\usepackage{amssymb}
\usepackage{pifont}
\usepackage{wrapfig}

\newcommand{\equationref}[1]{{Eq.~\ref{#1}}}
\newcommand{\tableref}[1]{{Table~\ref{#1}}}
\newcommand{\figref}[1]{{Fig.~\ref{#1}}}
\newcommand{\algorithmref}[1]{{Algorithm~\ref{#1}}}
\newcommand{\appendixref}[1]{{Appendix~\ref{#1}}}

\newcommand{\rom}[1]{\romannumeral #1\relax}

\usepackage[utf8]{inputenc} % allow utf-8 input
\usepackage[T1]{fontenc}    % use 8-bit T1 fonts
\usepackage{hyperref}       % hyperlinks
\usepackage{url}            % simple URL typesetting
\usepackage{booktabs}       % professional-quality tables
\usepackage{amsfonts}       % blackboard math symbols
\usepackage{nicefrac}       % compact symbols for 1/2, etc.
\usepackage{microtype}      % microtypography
\usepackage{xcolor}         % colors
\usepackage{multirow}
\usepackage{amsmath}
\usepackage{subcaption}
\usepackage{tcolorbox}
\tcbuselibrary{breakable}

\title{Pose-RFT: Enhancing MLLMs for 3D Pose Generation via Hybrid Action Reinforcement Fine-Tuning}

\author{%
  Bao Li\textsuperscript{{\normalfont 1,2}} \enspace
  Xiaomei Zhang\textsuperscript{{\normalfont 1,2}} \enspace
  Miao Xu\textsuperscript{{\normalfont 1,3}} \enspace
  Zhaoxin Fan\textsuperscript{{\normalfont 4}} \enspace
  Xiangyu Zhu\textsuperscript{{\normalfont 1,2}} \enspace
  Zhen Lei\textsuperscript{{\normalfont 1,2,3}} \\
  \textsuperscript{1}CASIA \quad
  \textsuperscript{2}UCAS \quad
  \textsuperscript{3}CAIR, HKISI, CAS \quad
  \textsuperscript{4}Beihang University
   \\
  \texttt{\{libao2023, zhangxiaomei2016, xiangyu.zhu, zhen.lei\}@ia.ac.cn} \\
  \texttt{miao.xu@cair-cas.org.hk} \quad
  \texttt{zhaoxinf@buaa.edu.cn}
}

\begin{document}

\maketitle

\begin{abstract}
Generating 3D human poses from multimodal inputs such as images or text requires models to capture both rich spatial and semantic correspondences. 
While pose-specific multimodal large language models (MLLMs) have shown promise in this task, they are typically trained with supervised objectives such as SMPL parameter regression or token-level prediction, which struggle to model the inherent ambiguity and achieve task-specific alignment required for accurate 3D pose generation. 
To address these limitations, we propose Pose-RFT, a reinforcement fine-tuning framework tailored for 3D human pose generation in MLLMs. We formulate the task as a hybrid action reinforcement learning problem that jointly optimizes discrete language prediction and continuous pose generation. 
To this end, we introduce HyGRPO, a hybrid reinforcement learning algorithm that performs group-wise reward normalization over sampled responses to guide joint optimization of discrete and continuous actions.
Pose-RFT further incorporates task-specific reward functions to guide optimization towards spatial alignment in image-to-pose generation and semantic consistency in text-to-pose generation.
Extensive experiments on multiple pose generation benchmarks demonstrate that Pose-RFT significantly improves performance over existing pose-specific MLLMs, validating the effectiveness of hybrid action reinforcement fine-tuning for 3D pose generation.
\end{abstract}

\section{Introduction}
% introduce pose-specific MLLMs
Recent advances in 3D human pose generation \cite{delmas2022posescript, delmas2024poseembroider,hong2022avatarclip,tevet2022human}  have increasingly focused on addressing the problem of understanding and reasoning about 3D human poses from multimodal inputs, such as images and text. Among these, pose-specific multimodal large language models (MLLMs) \citep{feng2024chatpose,li2024unipose} have emerged as a promising direction, extending general-purpose language models with dedicated pose decoders to enable joint reasoning over language, vision, and 3D pose. These models have shown strong performance across both image-to-pose and text-to-pose generation tasks.

\begin{figure}
    \centering
    \includegraphics[width=1.0\textwidth]{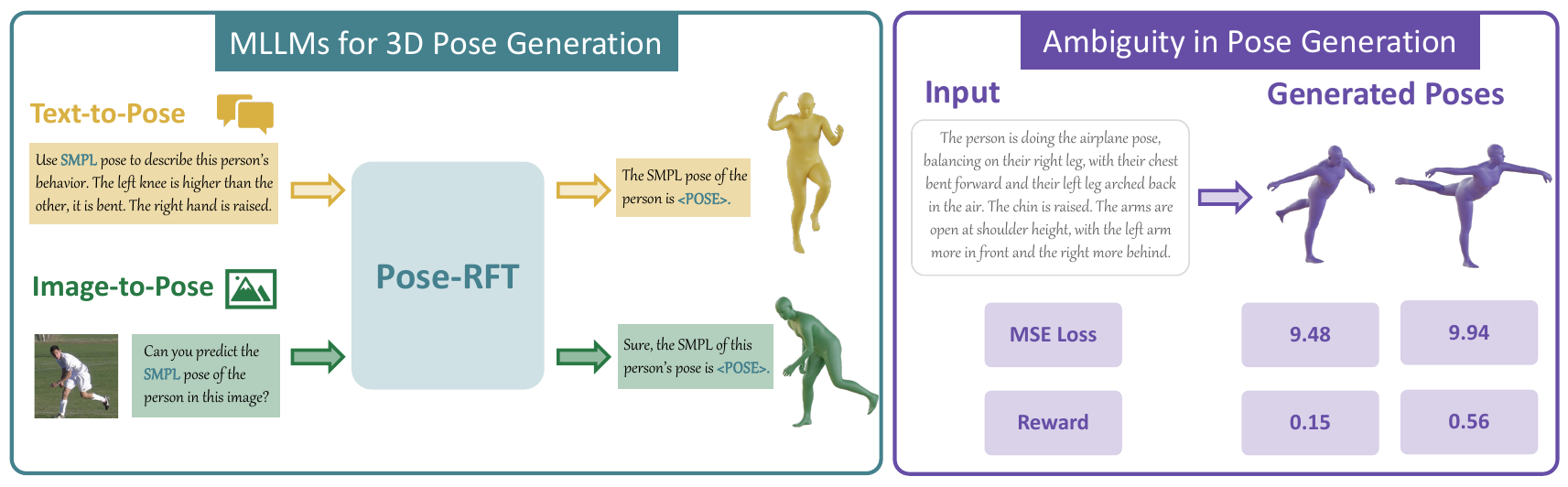}
    \vspace{-2.0 em}
    \caption{\textbf{Examples and Motivation.} \textbf{Left:} Our proposed Pose-RFT, an enhanced multimodal large language model, generates 3D SMPL human pose given either text or image input.  \textbf{Right:} Two 3D poses generated from the same prompt yield comparable MSE loss, while the reward reflects their difference in semantic alignment.}
    \label{fig1}
    \vspace{-1.5 em}
\end{figure}

% ambiguity and task-specific alignment
Despite their progress, current pose-specific MLLMs are typically trained with supervised objectives such as SMPL \citep{loper2023smpl} parameter regression \citep{feng2024chatpose} or token-level prediction \citep{li2024unipose}. While these objectives enforce consistency with annotated data, they are insufficient for capturing the inherent ambiguity and the task-specific spatial and semantic alignment required in 3D pose generation. 
In image-to-pose generation, ambiguity arises from depth and perspective limitations in 2D images, which often correspond to multiple plausible 3D poses. 
For text-to-pose generation, the challenge is further compounded by the vagueness and subjectivity of natural language, which results in a broad distribution over valid poses. As illustrated in \figref{fig1}, distinct 3D poses may yield similar reconstruction losses (e.g., MSE) despite exhibiting clear semantic differences. This highlights the limitations of standard supervised objectives and motivates the need for reward signals that better reflect semantic alignment.

These challenges motivate the use of reinforcement learning (RL), which provides a principled approach for optimizing models beyond supervised labels,  toward more aligned spatial and semantic outputs. However, most existing reinforcement fine-tuning (RFT) algorithms \citep{achiam2023gpt, jaech2024openai, liu2025visual, shen2025vlm, li2025videochat} operate in purely discrete token spaces and are primarily designed for language-level alignment. Such discrete approaches are inherently limited for optimizing the fine-grained, continuous outputs required in 3D human pose generation.

To address these challenges, we propose Pose-RFT (see \figref{fig2}), a reinforcement fine-tuning framework specifically designed for 3D human pose generation in MLLMs. \textbf{First}, we formulate the task as a hybrid action space reinforcement learning problem, where the policy simultaneously produces discrete actions (e.g., text tokens) and continuous actions (e.g., 3D pose parameters). 
Inspired by prior work in hybrid RL \citep{lowe2017multi,fan2019hybrid, li2021hyar}, we construct a unified representation space leveraging the language-aligned multimodal embeddings of existing pose-specific MLLMs. 
The continuous action is modeled by a multivariate Gaussian policy, parameterized by a pose head that outputs both the mean and covariance.
\textbf{Second}, we introduce HyGRPO, an online hybrid reinforcement learning algorithm built upon GRPO \citep{shao2024deepseekmath}, which directly optimizes policies over the original hybrid action space. For each input, the pretrained pose-specific MLLM generates multiple hybrid responses (text + pose), normalizes the reward scores within the group, and updates the policy to prefer responses with higher rewards. \textbf{Third}, we propose four verifiable, task-specific reward functions to guide policy optimization: (\rom{1}) a spatial location reward for image-to-pose generation, (\rom{2}) a semantic alignment reward for text-to-pose generation, (\rom{3}) a format correctness reward, and (\rom{4}) a text embedding similarity reward. By training with diverse outputs and structured feedback, HyGRPO encourages the model to generate 3D poses that are spatially accurate and semantically aligned.

In summary, our main contributions are as follows: 

(1) We propose Pose-RFT, the first reinforcement fine-tuning framework specifically designed for 3D human pose generation in MLLMs. (2) We develop HyGRPO, a hybrid-action reinforcement learning algorithm that effectively optimizes both discrete and continuous outputs in pose-specific MLLMs. (3) Extensive experiments on multiple pose generation benchmarks demonstrate that Pose-RFT significantly improves performance over existing pose-specific MLLMs, validating the effectiveness of hybrid action reinforcement fine-tuning for 3D pose generation.

\section{Related Work}
% referred from Visual-RFT, R1-VL
\subsection{Human Pose Generation}
Human pose generation involves producing 3D human poses conditioned on either images or text. 
For image-to-pose generation, also known as pose estimation, existing approaches are typically divided into optimization-based and regression-based methods. 
Optimization-based methods~\citep{bogo2016keep, pavlakos2019expressive} estimate 3D pose parameters by aligning projected joints with detected 2D keypoints through iterative refinement.
In contrast, regression-based approaches~\citep{hmrKanazawa17, cai2023smpler, dwivedi2024tokenhmr, goel2023humans} rely on deep neural networks to directly predict 3D poses from input images.
Text-to-pose generation aims to synthesize 3D human poses based on textual descriptions, such as physical attributes or actions \citep{delmas2022posescript, tevet2022human, hong2022avatarclip}. 
Although these methods have shown promising results, they remain confined to either image-to-pose or text-to-pose generation, without a unified framework capable of leveraging cross-modal knowledge to infer human poses from both visual and textual inputs.

\begin{figure}
    \centering
    \includegraphics[width=1.0\textwidth]{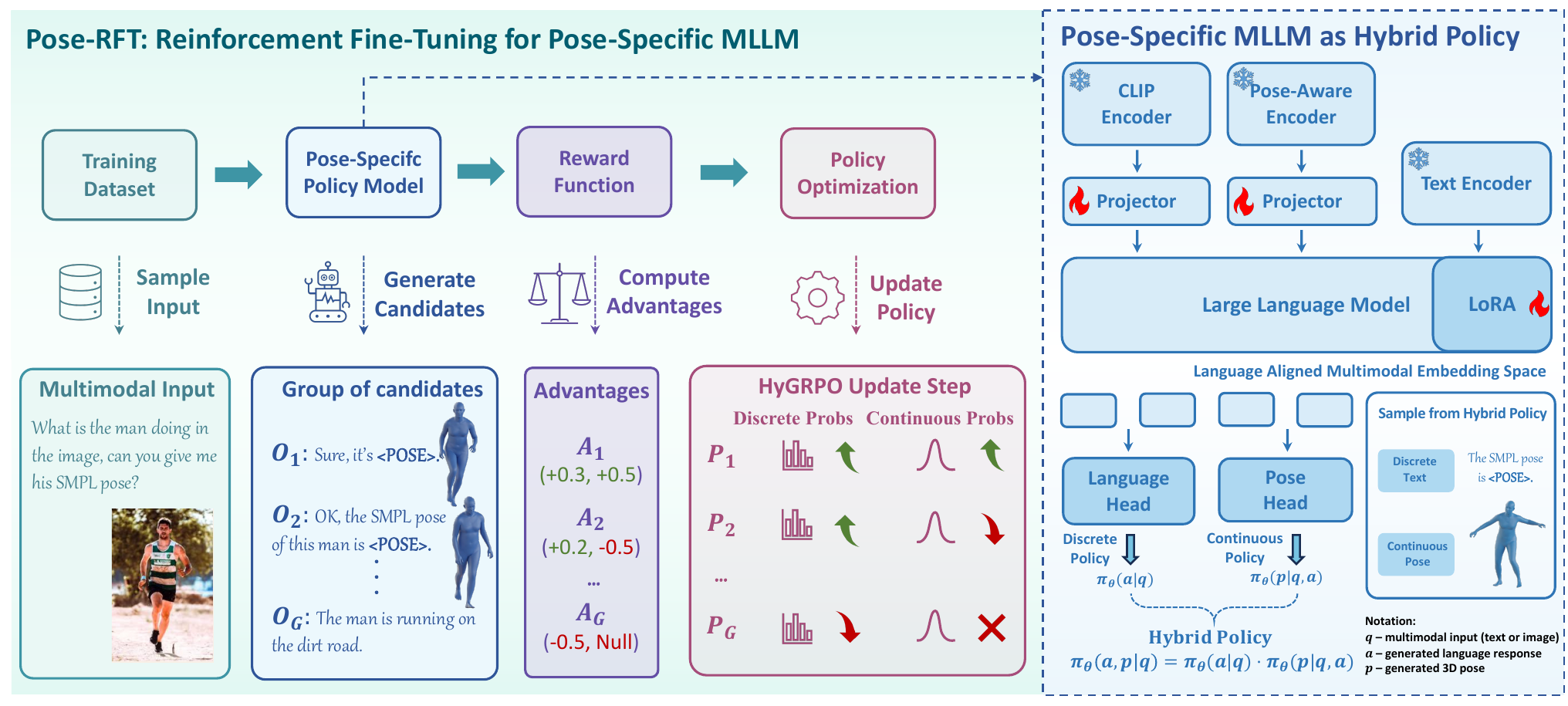}
    \caption{\textbf{Overview of Pose-RFT Framework.} A Reinforcement fine-tuning pipeline for pose-specific MLLMs. Given a multimodal input, the model generates multiple hybrid responses combining discrete text and continuous 3D pose outputs. Rewards are computed using task-specific metrics, normalized within each group, and used to update the model by promoting high-reward responses.}
    \label{fig2}
    \vspace{-2.0em}
  \end{figure}

\subsection{Multimodal Large Language Models}
Multimodal Large Language Models (MLLMs) \citep{achiam2023gpt,liu2023visual,li2024llava,wang2024qwen2,chen2024internvl} have shown strong performance in vision-language understanding tasks by jointly modeling visual inputs and natural language. These models excel at multimodal reasoning, visual grounding, and instruction following, enabling them to comprehend complex visual content in diverse application scenarios. Leveraging these capabilities, recent works have successfully applied MLLMs to downstream vision-centric tasks such as image segmentation \citep{lai2024lisa, bai2024one}, anomaly detection \citep{gu2024anomalygpt}, and keypoint localization \citep{wang2024locllm}, demonstrating their transferability beyond purely linguistic domains.

To adapt MLLMs to downstream tasks, post-training strategies such as supervised fine-tuning (SFT) and reinforcement fine-tuning (RFT) are commonly used. Recent efforts such as ChatPose \citep{feng2024chatpose} and UniPose \citep{li2024unipose} have applied SFT to extend MLLMs for 3D pose generation, leveraging their vision-language reasoning capabilities. However, these methods rely solely on SFT and do not incorporate reinforcement-based optimization. The absence of RFT limits the model’s capacity to further refine generation quality, particularly in scenarios involving ambiguity and task-specific alignment.

\subsection{Reinforcement Learning}
Reinforcement learning (RL) \citep{sutton1998reinforcement} is a core paradigm in machine learning, where an agent learns a policy-a mapping from observations to actions-by interacting with an environment and optimizing cumulative rewards. Through trial-and-error learning, the agent improves its policy based on feedback in the form of scalar rewards. Classical algorithms such as Q-learning\citep{watkins1992q} have been successfully applied in fields such as robotics, autonomous control, and game playing. With the rise of large language models \citep{radford2018improving, touvron2023llama, achiam2023gpt}, Reinforcement Learning with Human Feedback (RLHF) \citep{bai2022training} has become as a key technique for fine-tuning models using human preference data. RLHF leverages algorithms like Proximal Policy Optimization (PPO) \citep{schulman2017proximal} and Direct Preference Optimization (DPO) \citep{rafailov2023direct} to guide model behavior for improving the alignment, coherence, and helpfulness in response generation.

% works like OpenAI-o1 \citep{jaech2024openai} and DeepSeek-R1 \citep{guo2025deepseek} demonstrate the effectiveness of RL in improving performance on tasks involving code generation and mathematical reasoning.
In the context of multimodal large language models, recent works \citep{zhou2025r1, liu2025seg, zhan2025vision, liu2025visual, yang2025r1, zhang2025r1, shen2025vlm} have explored the use of RL with verifiable reward signals to enhance visual reasoning. However, the application of RL to 3D human pose generation remains underexplored, primarily due to the continuous nature of pose regression, which poses challenges for RL methods originally designed for discrete action spaces. To address similar challenges in other domains, several works have proposed hybrid discrete-continuous action formulations \citep{lowe2017multi, fan2019hybrid, li2021hyar}, offering a promising direction for adapting reinforcement learning to structured continuous tasks such as 3D pose generation.

\section{Methodology}
\label{Methodology}

This section first reformulates 3D human pose generation as a reinforcement learning problem under a hybrid action space. It then introduces the proposed \emph{Hybrid Action Space Group Relative Policy Optimization} (HyGRPO) algorithm, which jointly optimizes discrete language and continuous pose outputs. Finally, it describes how HyGRPO Finally, it describes how HyGRPO is applied to fine-tune pose-specific MLLMs using task-specific reward functions designed for 3D human pose generation.

\subsection{Reformulating 3D Pose Generation for Reinforcement Learning}
We formulate the 3D human pose generation within pose-specific MLLMs as a hybrid action reinforcement learning problem. The model operates in a hybrid action space comprising discrete language tokens and continuous 3D poses. The overall policy is defined as:
\begin{equation}
    \pi_{\theta}(a,p|q) = \pi_{\theta}(a|q) \cdot \pi_{\theta}(p|q,a), 
\end{equation}
where $q$ denotes a multimodal input, $a$ a represents discrete texual responses, and $p$ denotes continuous 3D pose parameters. We regard the joint distribution $\pi_{\theta}(a,p|q)$ as the overall policy, which is factorized into a discrete sub-policy $\pi_{\theta}(a|q)$ modeling the distribution over textual responses, and a continuous sub-policy $\pi_{\theta}(p|q,a)$ modeling the distribution over 3D poses conditioned on both the input query and the generated language response.

To parameterize the continuous policy, we adopt a multivariate Gaussian distribution defined over the space of 3D human poses:
\begin{equation}
\pi_{\theta}(p|q,a) = \mathcal{N}(p; \mu_{\theta}(q,a), \Sigma_{\theta}(q,a)),
\end{equation}
where the mean $\mu_{\theta}(q,a)$ and covariance $\Sigma_{\theta}(q,a)$ are predicted by a continuous pose head conditioned on the multimodal input $q$ and the discrete response $a$. This probabilistic formulation captures the aleatoric uncertainty inherent in 3D human pose generation by modeling the conditional distribution over continuous pose vectors. Furthermore, the use of a differentiable multivariate Gaussian enables both stochastic sampling during training and efficient gradient-based optimization within the continuous pose space.

Benefiting from the strong cross-modal alignment established during MLLM pretraining, both the discrete and continuous policies are built on a shared language-aligned multimodal embedding space. To further enhance pose understanding, we augment the original pose-specific MLLM with a pose-aware encoder module that enriches the multimodal state representation. Specifically, we incorporate a pose-specific Vision Transformer pretrained on human pose estimation tasks to extract high-resolution, pose-sensitive features.
These features are fused with the language-aligned multimodal embeddings to yield a more informative and pose-relevant state space for reinforcement learning. Details of the pose-aware encoder and the visual fusion strategy are provided in \appendixref{appendix1}. During optimization, each policy head is updated using its corresponding reward signal, while the shared backbone receives combined gradients, enabling end-to-end training across both discrete and continuous action spaces.

\subsection{HyGRPO: Hybrid Action Space Group Relative Policy Optimization}

We propose Hybrid Action Space Group Relative Policy Optimization (HyGRPO), an online reinforcement learning algorithm designed to enhance pose-specific MLLMs for 3D human pose generation. Leveraging the unified representation space learned during pretraining,
HyGRPO directly optimizes both the discrete language head and continuous pose head within the shared language-aligned multimodal embedding space,
facilitating coherent alignment between textual and pose outputs.
This algorithm effectively bridges the gap between discrete token prediction and continuous pose generation.

To handle hybrid outputs, HyGRPO models the policy $\pi_\theta$ over both discrete text answers $a$ and continuous human poses $p$ conditioned on input question $q$. For each training sample $q$ from dataset $\mathcal{D}$, we sample $G$ output candidates $\left\{a_i, p_i\right\}_{i=1}^G \sim \pi_\theta(\cdot | q)$, and optimize the policy using the following objective:
\begin{equation}
\begin{aligned}
\label{PPO}
\mathbb{E}_{q\sim\mathcal{D}, \left\{a_i, p_i\right\}_{i=1}^G \sim \pi_{\theta}(\cdot \mid q)} 
\left[
    \frac{1}{G} \sum_{i=1}^G r_i(\theta) \hat{A}_i
\right],
\end{aligned}
\end{equation}
where $r_i(\theta)$ is the importance weight of the $i$-th sampled output, computed as the ratio between the current policy and the reference policy:
\begin{equation}
\label{reward}
\begin{aligned}
    r_i(\theta) = \frac{\pi_{\theta}(a_i, p_i|q)}{\pi_\text{ref}(a_i,p_i|q)} = \underbrace{\frac{\pi_{\theta}(a_i|q)}{\pi_\text{ref}(a_i|q)}}_{r_d(a_i|q)} \cdot \underbrace{\frac{\pi_{\theta}(p_i|q,a_i)}{\pi_\text{ref}(p_i|q,a_i)}}_{r_c(p_i|q,a_i)} .
\end{aligned}
\end{equation}

We decompose the normalized advantage $\hat{A}$ into two components corresponding to the discrete and continuous actions:
\begin{equation}
\begin{aligned}
    \hat A(q,a,p) = \underbrace{\hat F(q,a)}_{\text{discrete advantages}} + \underbrace{\hat \Delta(q,a,p)}_{\text{continous advantages}},
\end{aligned}
\end{equation}
where $\hat F$ measures the quality of the generated textual response, and $\hat \Delta$ evaluates the predicted pose quality. To stabilize training, we adopt clipped importance sampling as in PPO \citep{schulman2017proximal}. The final HyGRPO training objective is:

\begin{equation}
\label{HyGRPO}
\begin{aligned}
\mathcal{J}_{\text{HyGRPO}} = 
& \mathbb{E}_{q\sim\mathcal{D}, \left\{a_i, p_i\right\}_{i=1}^G \sim \pi_{\theta}(\cdot \mid q)}
\Bigg[
\frac{1}{G} \sum_{i=1}^G \Big(
\min(r_d \hat{F}_i, \text{clip}(r_d, 1{-}\epsilon, 1{+}\epsilon)\hat{F}_i) \Big) \\
&+ \frac{1}{V} \sum_{i=1}^V \Big(\min(r_c \hat{\Delta}_i, \text{clip}(r_c, 1{-}\epsilon, 1{+}\epsilon)\hat{\Delta}_i) \Big)
- \beta D_{\text{KL}}(\pi_\theta \| \pi_{\text{ref}})
\Big) \Bigg],
\end{aligned}
\end{equation}

where $G$ is generated candidate group, $V$ is the candidate with pose output. 
This objective provides separate advantage signals for the discrete and continuous heads, while updating the shared backbone with combined gradients---enabling stable and generalizable training across the hybrid action space (see derivation in \appendixref{appendix2}). 

\algorithmref{algorithm_hygrpo} describes the pseudo-code of HyGRPO. At each iteration, we sample a mini-batch of questions, generate multiple candidate outputs from the current policy, compute both types of rewards using task-specific reward models, and update the policy using the HyGRPO objective.

\begin{algorithm}
\caption{\textbf{Hybrid Action Space Group Relative Policy Optimization}}
\label{algorithm_hygrpo}
\textbf{Input:} Initial policy model $\pi_{\text{init}}$ (a pretrained pose-specifc MLLM); reward models $R _\varphi$; \\ 
task dataset $\mathcal{D}$; hyperparameters $\epsilon$
\begin{algorithmic}
\State Initialize $\pi_\theta \gets \pi_{\text{init}}, \pi_{\text{ref}} \gets \pi_{\text{init}}$
\For{iteration $= 1, \dots, N$}
    \State Sample a batch $\mathcal{D}_b$ from $\mathcal{D}$
    \State Generate $G$ outputs $\{a_i,p_i\}_{i=1}^{G} \sim \pi_\theta(\cdot \mid q)$ for each question $q \in \mathcal{D}_b$
    \State Compute rewards $\{\mathcal{R}_i\}_{i=1}^{G}$ for each $(a_i, p_i)$ by running $\mathcal{R}_\varphi$
    \State Compute $\hat{A}_{i}$ for $(a_i, p_i)$ via group relative advantage estimation
    \State Update $\pi_\theta$ by maximizing HyGRPO objective \equationref{HyGRPO}
\EndFor
\textbf{Output:} policy model $\pi_\theta$
\end{algorithmic}
\end{algorithm}

\subsection{Enhance Pose-Specific MLLM with HyGRPO}
We apply HyGRPO to optimize pose-specific MLLMs over hybrid action outputs using task-specific reward signals. As illustrated in \figref{fig2}, for each question $q \in \mathcal{D}_b$, the policy model $\pi_{\theta}$ generates a group of $G$ hybrid outputs, each consisting of a discrete textual answer and a continuous 3D pose, i.e., $\left\{O_i\right\}_{i=1}^{G}$. For each candidate, we compute task-specific rewards that are carefully designed for different pose generation settings to guide policy updates through HyGRPO.

\paragraph{Joint Location Reward in Image-to-Pose Generation.}
In the image-to-pose generation task, the model is expected to output SMPL pose coefficients conditioned on the input image. 
To encourage spatial accuracy, the reward should reflect how well the predicted pose aligns with the visual input. 
A widely adopted metric in 3D human pose estimation is the mean joint position error, which computes the average Euclidean distance between predicted and ground-truth 3D joint locations. Following this, we define the reward as the inverse of the joint error, assigning higher scores to more accurate spatial alignment:
\begin{equation}
    \mathcal{R}_{\text{joint}} = \frac{1}{|| J_\text{pred} - J_\text{gt}||_2}.
\end{equation}

\paragraph{Semantic Alignment Reward in Text-to-Pose Generation.}
In the text-to-pose generation task, the model is expected to predict SMPL pose coefficients conditioned on a text prompt. Unlike image-to-pose generation, which emphasizes joint-level accuracy, this task focuses on high-level semantic alignment between the textual description and the generated pose.

To quantify this alignment, we adopt a pretrained text-pose retrieval model that maps both inputs into a shared embedding space. Specifically, the retrieval model comprises a text encoder $\phi_t(\cdot)$ and a pose encoder $\phi_p(\cdot)$, both projecting their respective inputs into a shared embedding space. The semantic alignment reward is defined as the similarity score between the encoded text and the generated pose: 
\begin{equation} \mathcal{R}_{\text{semantic}} = \text{cos}(\phi_t(q), \phi_p(p)) .
\end{equation}

\paragraph{Format Reward.}
To encourage the model to generate responses that conform to a specified format, we introduce a format reward, denoted as $R_{\text{format}}$. For instance, we expect the model to produce outputs enclosed in a template such as: \textit{“The SMPL pose of this person is <POSE>.”} To enforce this constraint, we apply regular expression matching to assess format compliance. The format reward is defined as:

\begin{equation}
\mathcal{R}_{\text{format}} =
\begin{cases}
1, & \text{if the output matches the expected format} \\
0, & \text{otherwise}
\end{cases}.
\end{equation}

\paragraph{Text Embedding Similarity Reward.}
To preserve general QA capabilities while fine-tuning for pose-centric tasks, we incorporate a text reward that encourages semantic agreement between generated and ground-truth answers in vision-language QA tasks. Specifically, we utilize the BGE-M3 encoder \citep{chen2024bge} to compute dense embeddings for both the model-generated answer and the ground-truth response. The reward is defined as the cosine similarity between the normalized embeddings of the predicted and ground-truth answers:
\begin{equation}
    \mathcal{R}_{\text{text}} = \cos(E(a_\text{pred}), E(a_\text{gt})).
\end{equation}

\section{Experiments}
\label{Experiments}
% Pretrain + RL

\subsection{Experimental Setup}
\label{Experimental Setup}
\paragraph{Datasets.}
To train Pose-RFT, we incorporate four types of data sources to enhance multimodal understanding: \textbf{(1) Text-Pose Data.} We utilize the PoseScript dataset \citep{delmas2022posescript}, which provides natural language descriptions paired with 3D human poses. This enables the model to learn fine-grained semantic correlations between language and human poses. \textbf{(2) Image-Pose Data.} Following prior works \citep{goel2023humans, feng2024chatpose, li2024unipose}, we adopt standard human pose estimation training datasets, including Human3.6M \citep{ionescu2013human3}, MPI-INF-3DHP \citep{mehta2017monocular}, COCO \citep{lin2014microsoft}, and MPII \citep{andriluka20142d}. For evaluation, we use the 3DPW \citep{von2018recovering} and Human3.6M test sets. \textbf{(3) Image-Text Data.} We employ the BEDLAM-Script dataset introduced in PoseEmbroider \citep{delmas2024poseembroider}, a curated multimodal dataset containing images, 3D poses, and textual descriptions, constructed based on the BEDLAM dataset \citep{black2023bedlam}. \textbf{(4) VQA Data.} For visual question answering, we utilize the LLaVA-Instruct-150k dataset \citep{liu2023visual}.

\begin{table*}
\centering
\caption{\textbf{Comparison on Human Pose Estimation task.} Reconstruction metrics are reported on the 3DPW and Human3.6M datasets.}
\begin{tabular}{l|cc|cc}
\toprule
\multirow{2}{*}{Method} & \multicolumn{2}{c|}{3DPW \citep{von2018recovering}} & \multicolumn{2}{c}{Human3.6M \citep{ionescu2013human3}} \\
\cmidrule(lr){2-3} \cmidrule(lr){4-5}
& MPJPE $\downarrow$ & PA-MPJPE $\downarrow$ & MPJPE $\downarrow$ & PA-MPJPE $\downarrow$ \\
\midrule
HMR \citep{von2018recovering} & 130.0 & 76.7 & 88.0 & 56.8 \\ 
PyMAF \citep{zhang2021pymaf} & 92.8 & 58.9 & 57.7 & 40.5 \\
    SMPLer \citep{xu2023smpler} & 73.7 & 43.4 & 45.2 & \textbf{32.4} \\
    HMR2.0 \citep{goel2023humans} & 70.0 & 44.5 & \textbf{44.8} & 33.6 \\
    Zolly \citep{wang2023zolly} & 76.2 & 47.9 & 49.4 & 32.3 \\
    MEGA \citep{fiche2024mega} & \textbf{67.5} & \textbf{41.0} & - & - \\
    TokenHMR \citep{dwivedi2024tokenhmr} & 71.0 & 44.3 & - & - \\
    \midrule
    ChatPose \citep{feng2024chatpose} & 163.6 & 81.9 & 126.0 & 82.4 \\
    %UniPose$^\dagger$ & 97.4 & 61.2 & \textbf{65.8} & \textbf{39.4} \\
    UniPose \citep{li2024unipose} & 94.7 & 59.1 & 69.2 & \textbf{41.8} \\
    Pose-RFT (Ours) & \textbf{85.9} & \textbf{51.6} & \textbf{63.0} & 44.5 \\
\bottomrule
\end{tabular}
\label{tab1}
\vspace{-1.0 em}
\end{table*}

\paragraph{Metrics.}
We evaluate our model on both image-to-pose and text-to-pose tasks using reconstruction and retrieval metrics. \textbf{Image-to-Pose Reconstruction Metrics:} We report the Mean Per Joint Position Error (MPJPE) and the Procrustes-aligned MPJPE (PA-MPJPE), which measure the average Euclidean distance between predicted and ground-truth joint positions, with and without Procrustes alignment, respectively. \textbf{Text-to-Pose Retrieval Metrics:} Following \citep{feng2024chatpose,li2024unipose}, we report Recall@K (K = 5, 10, 20) for both text-to-pose ($R^{T2P}$) and pose-to-text ($R^{P2T}$) retrieval tasks, which assess the accuracy of matching poses with their corresponding textual descriptions.

\paragraph{Implementation Details.} We adopt LLaVA-1.5V-7B \citep{liu2023visual} as the vision-language backbone. For the pose-aware encoder, we employ the pretrained Vision Transformer from \citep{goel2023humans}. Reinforcement fine-tuning follows the settings of Visual-RFT \citep{liu2025visual} and VLM-R1 \citep{shen2025vlm}. During both pretraining and fine-tuning, the CLIP encoder and the pose-aware encoder are kept frozen, while the projector and task head are updated. The large language model is fine-tuned using LoRA \citep{hu2022lora}. Further implementation details are provided in the Appendix \ref{appendix3}.

\subsection{Comparisons on Human Pose Generation Tasks}
\paragraph{Image-to-Pose Generation.} For the image-to-pose generation task, we compare Pose-RFT with traditional pose estimation approaches \citep{von2018recovering, zhang2021pymaf, xu2023smpler, goel2023humans, wang2023zolly, fiche2024mega, dwivedi2024tokenhmr} and MLLM-based approaches \citep{feng2024chatpose, li2024unipose}, on 3DPW \citep{von2018recovering} and Human3.6M \citep{ionescu2013human3} datasets. As shown in \tableref{tab1}, Pose-RFT significantly outperforms other pose-specific MLLMs, narrowing the performance gap between general-purpose MLLMs and task-specific pose estimation models. We attribute this improvement primarily to the pose-aware encoder, which captures more comprehensive pose-relevant information and effectively enhances the model’s ability to understand and generate human poses.

\begin{table*}
%\scriptsize\setlength{\tabcolsep}{7pt}
\setlength{\tabcolsep}{4pt}
\centering
\caption{\textbf{Comparison on Text-to-Pose Generation Task.} Retrieval metrics (Recall@K, K=5, 10, 20) are reported on the PoseScript dataset under two evaluation protocols.}
\begin{tabular}{l|ccc|ccc|ccc|ccc}
\toprule
\multirow{2}{*}{Method} 
& \multicolumn{6}{c|}{PoseScript (Full Retrieval)} & \multicolumn{6}{c}{PoseScript (Random Sampling)} \\
\cmidrule(lr){2-7} \cmidrule(lr){8-13}
& \multicolumn{3}{c|}{$\text{R}^{\text{T2P}} \uparrow$} & \multicolumn{3}{c|}{$\text{R}^{\text{P2T}} \uparrow$} 
& \multicolumn{3}{c|}{$\text{R}^{\text{T2P}} \uparrow$} & \multicolumn{3}{c}{$\text{R}^{\text{P2T}} \uparrow$} \\
\midrule
PoseScript \citep{delmas2022posescript} & 40.4 & 52.3 & 65.0 & 41.4 & 54.1 & 65.9 & 73.3 & 82.5 & 89.4 & 70.0 & 82.5 & 87.4  \\
ChatPose \citep{feng2024chatpose}       & 17.6 & 25.3 & 35.8 & 28.0 & 39.0 & 54.4 & 39.9 & 50.6 & 58.7 & 56.1 & 65.3 & 72.5 \\
ChatHuman \citep{lin2024chathuman}      & 41.8 & 52.6 & 65.1 & 42.1 & 52.3 & 66.5 & - & - & - & - & - & - \\
UniPose \citep{li2024unipose}           & - & - & - & - & - & - & \textbf{73.7} & 82.4 & \textbf{89.6} & 70.9 & 80.5 & 89.6 \\
\midrule
Pose-RFT (Ours)                         & \textbf{42.2} & \textbf{53.0} & \textbf{65.5} & \textbf{45.3} & \textbf{57.2} & \textbf{70.4} & 71.8 & \textbf{82.6} & 88.7 & \textbf{74.6} & \textbf{86.5} & \textbf{91.5} \\
\bottomrule
\end{tabular}
\label{tab2}
\vspace{-1.6 em}
\end{table*}

\paragraph{Text-to-Pose Generation.} For the text-to-pose generation task, we compare Pose-RFT with existing text-conditional pose generation models \citep{delmas2022posescript, feng2024chatpose, lin2024chathuman, li2024unipose} on PoseScript-H2 test set \citep{delmas2022posescript}, which contains 1234 high-quality human-written text-pose pairs. Since these generation models do not natively support retrieval, following previous works, we generate 3D poses from the input captions and then evaluate the results using a pretrained text-to-pose retrieval model provided by \citep{delmas2022posescript}, computing Recall@K scores as a proxy for generation quality. As shown in \tableref{tab2}, Pose-RFT achieves the best performance across most metrics. We attribute this primarily to the reinforcement fine-tuning with a semantic alignment reward,  which effectively enhances the model's ability to capture fine-grained text-pose correspondence and improves generation quality. 

While prior work on text-to-pose generation is limited, existing methods evaluated their performance on the PoseScript-H2 benchmark using two different retrieval protocols. To ensure fair and comprehensive comparison, we report results under both protocols: \textbf{Protocol 1 (Full Retrieval)} follows the standard one-to-one retrieval setting, where each query corresponds to a single target in the entire test set. We compute Recall@K over the full test set, where each query is matched to its corresponding ground-truth pose based on dataset pairing.
\textbf{Protocol 2 (Random Sampling)}, in contrast, adopts the commonly used Recall@K evaluation under random sampling. For each query, we randomly sample $N=32$ candidates (including the ground-truth), and repeat the retrieval process $R=10$ times to average out the variance.

\begin{figure}
  \centering
  \begin{subfigure}
  {0.49\textwidth}
    \centering
    \includegraphics[width=\linewidth]{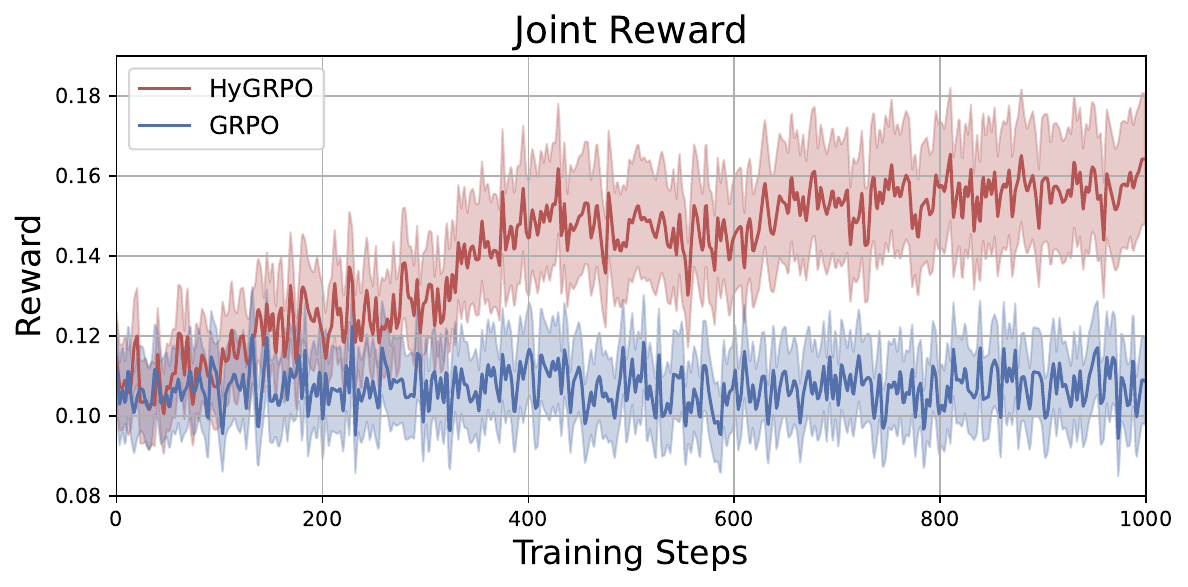}
    \label{fig:2a}
  \end{subfigure}%
  \hfill
  \begin{subfigure}
  {0.49\textwidth}
    \centering
    \includegraphics[width=\linewidth]{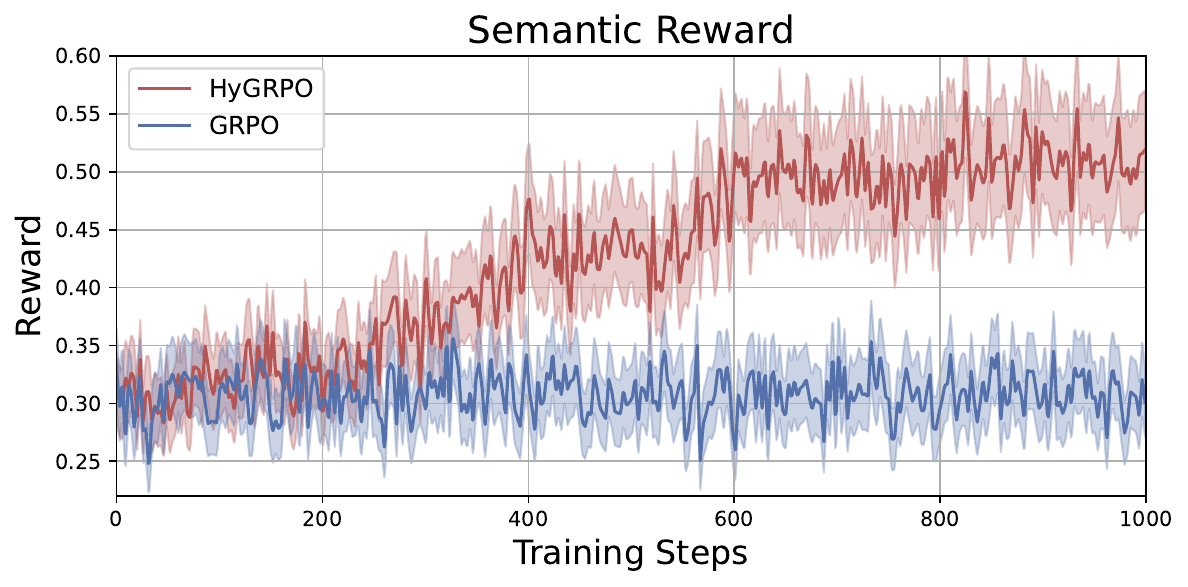}
    \label{fig:2b}
  \end{subfigure}
\vspace{-1.5 em}
\caption{\textbf{Comparison between GRPO and HyGRPO.} The horizontal axis represents the number of training steps. The vertical axis shows the mean group reward, computed across sampled candidate poses for each input.}
\label{fig3}
\vspace{-2.0 em}
\end{figure}

\subsection{Ablation Studies and Discussions}
\begin{wrapfigure}{r}{0.5\textwidth}  % "r" 表示图在右侧, 0.5\textwidth 是宽度
    \centering
    \includegraphics[width=0.48\textwidth]{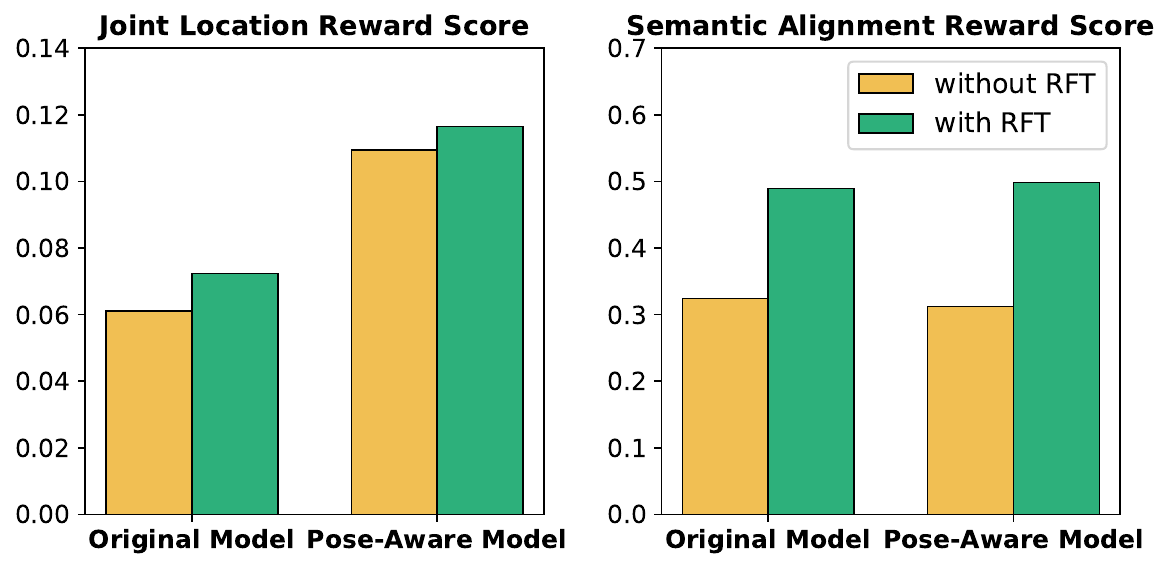}
    \caption{Ablation study of Pose-Aware Encoder (left/right bars in each group) and Reinforcement Fine-tuning (color-coded). Semantic alignment reward score are averaged over the PoseScript-H2 test set, while joint location reward scores are averaged over the 3DPW test set.}
    \label{fig4}
  \end{wrapfigure}

\paragraph{Pose-Aware Encoder.} We evaluate the effectiveness of the proposed pose-aware encoder across the full training process. As shown in \figref{fig4}, the pose-aware model achieves a higher joint location reward score on the 3DPW dataset compared to the baseline model. This result indicates that relying solely on a CLIP encoder is suboptimal for pose estimation tasks. In contrast, by incorporating an additional Vision Transformer pretrained on pose estimation and adopting a token-level feature fusion strategy, the pose-aware model captures fine-grained pose information more effectively, leading to significant improvements in pose accuracy. However, we observe that the pose-aware encoder brings little benefit to the text-to-pose semantic reward, possibly because the additional visual tokens are less aligned with the textual input, limiting their contribution to cross-modal understanding.

\paragraph{GRPO vs. HyGRPO.} We compare the effectiveness of GRPO and the proposed HyGRPO algorithm for reinforcement fine-tuning of pose-specific MLLMs. In this experiment, both methods are applied to models trained on the multi-task dataset. As shown in \figref{fig3}, GRPO, which solely optimizes next-token prediction in the discrete action space, brings negligible improvement to the quality of 3D pose generation. In contrast, HyGRPO jointly optimizes discrete token outputs and continuous 3D pose parameters under the guidance of task-specific reward functions. Leveraging the semantic alignment and joint location reward---both of which quantitatively assess output quality---HyGRPO yields consistent improvements in reward scores during training, indicating better alignment in both text- and image-conditioned pose generation.

\paragraph{Reinforcement Fine-tuning.} We evaluate the effectiveness of Pose-RFT through reinforcement fine-tuning using HyGRPO. Starting from a pretrained pose-specific MLLM, the model is fine-tuned on the multi-task dataset for 1,000 steps. As shown in \figref{fig4}, reinforcement fine-tuning consistently improves both text-to-pose and image-to-pose generation performance on the PoseScript and 3DPW datasets, respectively, with particularly notable gains in the text-to-pose setting. These results demonstrate the efficacy of reinforcement learning in enhancing the alignment between language, vision, and 3D pose representations. Additionally, we observe that the reward improvement on the text-to-pose task is substantially greater than that on the image-to-pose task. We attribute this to the strong joint modeling capability of the pretrained text-pose retrieval model \citep{delmas2022posescript}, that effectively guide text-conditioned pose generation.

\paragraph{Distributional Modeling in Pose Generation}
To assess the impact of modeling 3D human pose generation as a probabilistic distribution, we conduct an ablation study comparing our Gaussian policy to a standard deterministic regression baseline. As shown in \tableref{tab3}, without reinforcement fine-tuning, the distributional head performs slightly worse than the deterministic counterpart. However, when reinforcement signals are introduced, the probabilistic model achieves significantly better performance. These results suggest that distributional modeling facilitates more effective reward-driven learning, enabling the model to better integrate semantic and spatial feedback from multimodal inputs and produce higher-quality 3D poses.

\begin{table}
\setlength{\tabcolsep}{4pt}
\caption{Ablation study on distributional modeling (denoted as “Dist.”) for 3D pose generation. Reconstruction and retrieval metrics are reported on the 3DPW and PoseScript-H2 datasets.}
\centering
\begin{tabular}{c|cc|cc|cc}
\toprule
\multicolumn{1}{c|}{Method} & \multicolumn{1}{c}{Dist.} & \multicolumn{1}{c|}{RFT} & \multicolumn{2}{c|}{\textbf{Image-to-Pose Generation }} & \multicolumn{2}{c}{\textbf{Text-to-Pose Generation}} \\
 & & & MPJPE $\downarrow$ & PA-MPJPE $\downarrow$ & $\text{mRecall}^{\text{T2P}} \uparrow$ & $\text{mRecall}^{\text{P2T}} \uparrow$ \\
\midrule
Baseline &  \ding{55} & \ding{55} & 90.4 & 57.1 & 36.2 & 41.5 \\
Baseline + Dist. & \ding{51} & \ding{55} & 91.4 & 59.2 & 37.4 & 42.0 \\
Baseline + Dist. + RFT & \ding{51} & \ding{51} & 85.9 & 51.6 & 53.6 & 57.6 \\
\bottomrule
\end{tabular}
\label{tab3}
\vspace{-1.0 em} 
\end{table}

\begin{figure}
  \centering
  \includegraphics[width=1.0\textwidth]{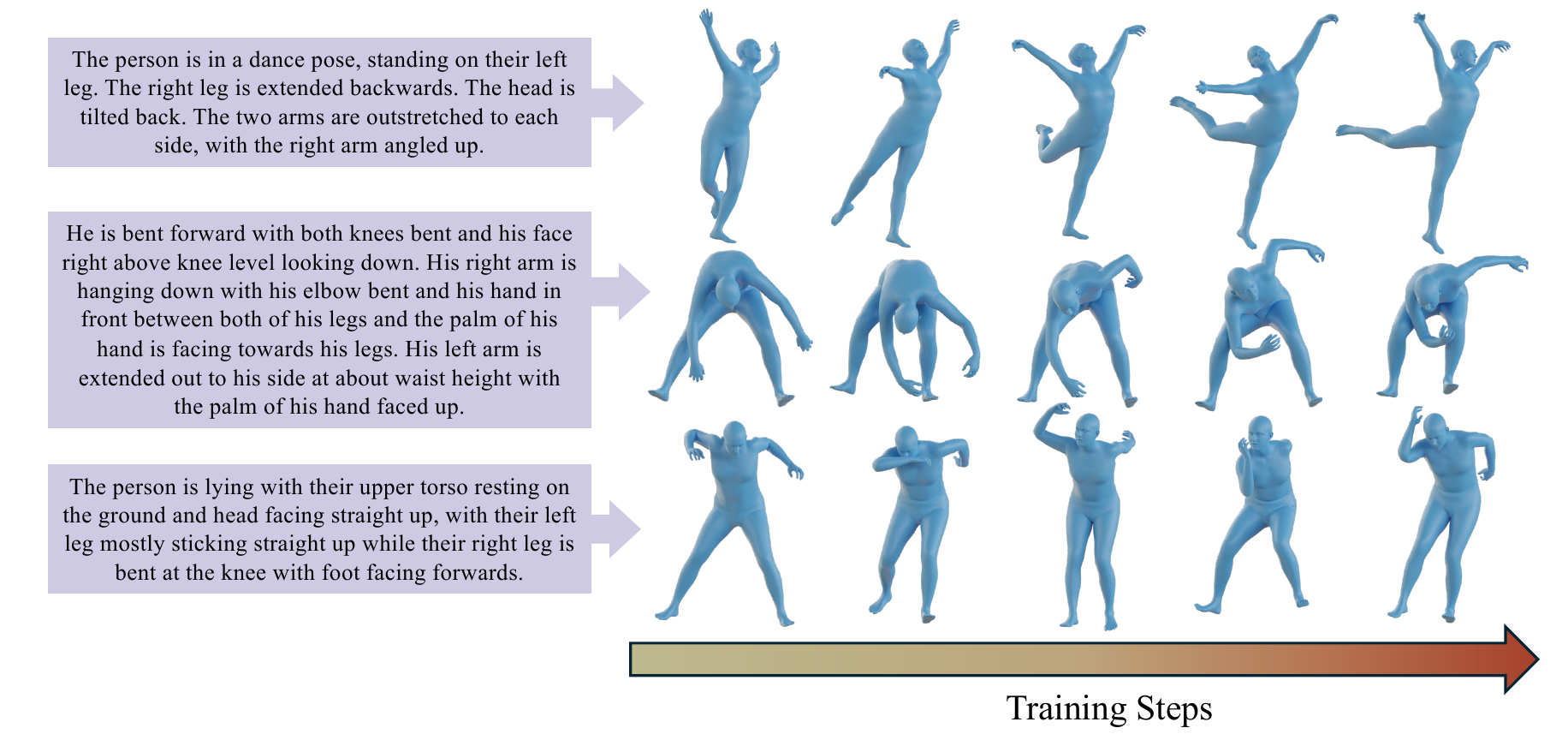}
  \caption{\textbf{Training progression of text-to-pose generation.}
  3D poses sampled at different reinforcement fine-tuning steps, conditioned on the same text prompt. As reinforcement fine-tuning progresses with the semantic alignment reward, the generated poses exhibit increasingly improved semantic consistency and structural plausibility.}
  \label{fig5}
  \vspace{-1.0 em} 
\end{figure}

\subsection{Qualitative Results}
As shown in \figref{fig5}, we present qualitative examples of 3D poses generated from a fixed set of text prompts. For each prompt, we visualize poses sampled at different reinforcement fine-tuning steps (left to right). These results illustrate the model’s progressive improvement in generating poses that are semantically aligned with the corresponding textual descriptions.

\section{Conclusion}
In this paper, we present Pose-RFT, the first reinforcement fine-tuning framework specifically designed to enhance 3D pose generation in pose-specific MLLMs. To address the challenges of discrete-continuous hybrid action space, we introduce Hybrid Action Space Group Relative Policy Optimization (HyGRPO), a novel reinforcement learning algorithm that jointly optimizes textual responses and 3D pose outputs. By incorporating task-specific reward functions that capture both spatial accuracy and semantic alignment, Pose-RFT effectively improves generation quality under different input modalities. Extensive experiments across multiple benchmarks demonstrate that Pose-RFT consistently outperforms existing pose-specific MLLMs, validating the effectiveness of hybrid action reinforcement fine-tuning for 3D pose generation tasks.

\bibliographystyle{plain}
\bibliography{reference}

\appendix

\section{Details of Pose-Aware Encoder and the Visual Fusion Strategy}
\label{appendix1}
This section provides additional implementation details of the Pose-Aware Encoder and the corresponding visual feature fusion strategy used in our framework. 

Previous works \citep{liu2023visual, feng2024chatpose} typically adopt the CLIP visual encoder \citep{radford2021learning} as the visual backbone. However, since CLIP is pretrained using global and coarse-grained supervision from image-caption pairs, it struggles to capture fine-grained pose details. In contrast, pose estimation tasks require precise localization of human keypoints, encouraging the encoder to learn fine-grained, pose-aware representations. To address this limitation, we introduce a pose-specific Vision Transformer \citep{goel2023humans} pretrained on human pose estimation into the visual pipeline, as illustrated in \figref{fig1}.

Let $f_a$ denote the CLIP visual encoder and $f_b$ the pose-aware encoder. Given an input image $x$, we extract two sets of visual embeddings: $v_a = f_a(x) \in \mathbb{R}^{L_v \times d_a}$ and $v_b = f_b(x) \in \mathbb{R}^{L_v \times d_b}$, where $L_v$ is the number of visual tokens, and $d_a$, $d_b$ are the respective embedding dimensions. While UniPose \citep{li2024unipose} directly concatenates $v_a$ and $v_b$ along the channel dimension and applies a single linear projection $W \in \mathbb{R}^{(d_a + d_b) \times d}$
this design can lead to patch-level misalignment due to the differing preprocessing pipelines of the two encoders, potentially degrading performance.

To preserve the individual strengths of each visual encoder, we propose to project $v_a$ and $v_b$ individually using two separate linear layers: 
\begin{equation}
    v_a' = W_a v_a, \quad v_b' = W_b v_b, \quad W_a \in \mathbb{R}^{d_a \times d}, \quad W_b \in \mathbb{R}^{d_b \times d}
\end{equation}
The transformed features $v_a'$ and $v_b'$ are then used in a token-level fusion with language features during joint training. This design maintains the representational integrity of both visual encoders while aligning their output with the language model’s embedding space.

\section{Theoretical Derivation of the HyGRPO Objective}
\label{appendix2}
Our goal is to optimize a hybrid policy $\pi_{\theta}(a,p|q)$, where $a$ is a discrete action (e.g., text sequence), and $p$ is continuous action (e.g., 3D human pose), both conditioned on the input $q$. We assume the policy factorizes as:
\begin{equation}
    \pi_{\theta}(a,p|q) = \pi_{d}(a|q) \cdot \pi_{c}(p|q,a),
\end{equation}
where $\pi_d$ is the discrete policy and $\pi_c$ is the continuous policy conditioned on the discrete output. To simplify the derivation, we temporarily exclude the clipping and KL regularization terms from the GRPO \citep{shao2024deepseekmath} objective. These components are included in the final training objective but are omitted here for clarity. We begin with the simplified form of the GRPO objective :
\begin{equation}
\begin{aligned}
\mathbb{E}_{q\sim\mathcal{D}, \left\{a_i, p_i\right\}_{i=1}^G \sim \pi_{\theta}(\cdot \mid q)}
\Bigg[
\frac{1}{G} \sum_{i=1}^G 
r_i(\theta) \hat{A}_i  \Bigg].
\end{aligned}
\end{equation}

Here, $r_i(\theta)$ is the importance weight of the $i$-th sampled output, computed as the ratio between the current policy and the reference policy:
\begin{equation}
\begin{aligned}
    r_i(\theta) = \frac{\pi_{\theta}(a_i, p_i|q)}{\pi_\text{ref}(a_i,p_i|q)} = \underbrace{\frac{\pi_{\theta}(a_i|q)}{\pi_\text{ref}(a_i|q)}}_{r_d(a_i|q)} \cdot \underbrace{\frac{\pi_{\theta}(p_i|q,a_i)}{\pi_\text{ref}(p_i|q,a_i)}}_{r_c(p_i|q,a_i)} .
\end{aligned}
\end{equation}

To effectively train the hybrid policy, we decompose the surrogate loss into discrete and continuous components. This is motivated by the nature of our task design, where the rewards are defined separately for the discrete and continuous outputs: textual rewards $R_d(q,a)$ evaluate the semantic correctness of the generated answer $a$. pose rewards $R_c(q,a,p)$  measures the plausibility and relevance of the generated pose $p$ conditioned on both the question and answer. 

Accordingly, we decompose the advantage estimate into discrete and continuous components:
\begin{equation}
\begin{aligned}
    \hat A(q,a,p) = \underbrace{\hat F(q,a)}_{\text{discrete advantages}} + \underbrace{\hat \Delta(q,a,p)}_{\text{continous advantages}}.
\end{aligned}
\end{equation}

This decomposition does not rely on an additive assumption over a shared reward function. Instead, it reflects the fact that the discrete and continuous components are supervised by independent reward signals tailored to their modalities. Accordingly, we compute two independent advantages from these separate rewards, using per-sample normalization within the candidate set:
\begin{equation}
    \hat F(q,a_i) = \frac{R_d^{(i)} - \text{mean}(\{R_d\}_{i=1}^G)}{\text{std}(\{R_d\}_{i=1}^G)} \qquad 
    \hat \Delta_i(q,a_i,p_i) = \frac{R_c^{(i)} - \text{mean}(\{R_c\}_{i=1}^G)}{\text{std}(\{R_c\}_{i=1}^G)}.
\end{equation}

We now substitute this decomposition into the GRPO objective. To facilitate this, we first move the expectation over the continuous action into an inner term:

\begin{equation}
\begin{aligned}
    \mathbb{E}_{q \sim \mathcal{D}, 
    \left\{a_i\right\}_{i=1}^G \sim \pi_{\theta}(\cdot \mid q)} 
    \Bigg[ \frac{1}{G} \sum_{i=1}^{G} r_d(a_i|q) \underbrace{\mathbb{E}_{p_i \sim \pi_{\theta}(p|q,a_i)} [r_c(p_i|q,a_i) \hat A_i(q,a_i,p_i)]}_{=:G(q,a_i)} \Bigg],
\end{aligned}
\end{equation}

We then analyze the inner term $G(q,a_i)$ by substituting the advantage decomposition:
\begin{equation}
\begin{aligned}
    G(q,a_i) &= \mathbb{E}_{p_i \sim \pi_{\theta}(p|q,a_i)}[r_c(p_i|q,a_i) \hat F_i(q,a_i) + r_c(p_i|q,a_i) \hat \Delta_i(q,a_i,p_i)] \\
    &= \hat F(q,a_i) \underbrace{\mathbb{E}_{p_i \sim \pi_{\theta}(p|q,a_i)}[r_c(q,a_i,p_i)]}_{\text{=1}} +  \mathbb{E}_{p_i \sim \pi_{\theta}(p|q,a_i)} [r_c(q,a_i,p_i) \hat \Delta_i(q,a_i,p_i)] \\
    &= \hat F_i(q,a_i) + \mathbb{E}_{p_i \sim \pi_{\theta}(p|q,a_i)} [r_c(q,a_i,p_i) \hat \Delta_i(q,a_i,p_i)].
\end{aligned}
\end{equation}

Substituting $G(q,a_i)$ back into the outer expectation, we arrive at a natural decomposition into two components:
\begin{equation}
\begin{aligned}
 \underbrace{\mathbb{E}_{q \sim \mathcal{D}, 
\left\{a_i\right\}_{i=1}^G \sim \pi_{\theta}(\cdot \mid q)} 
 [r_d(a_i|q)\hat F_i(q,a_i) ]}_{\mathcal{J}_d} 
 + 
 \underbrace{\mathbb{E}_{q \sim \mathcal{D}, \left\{a_i, p_i\right\}_{i=1}^G \sim \pi_{\theta}(\cdot \mid q) }[r_d(q,a_i)r_c(q,a_i,p_i)\hat \Delta_i (q,a_i,p_i)]}_{\mathcal{J}_c}.
\end{aligned}
\end{equation}

Although the discrete importance weight $r_d(a|q)$ and the continuous policy $\pi_{\theta}(p|q,a)$ share a common embedding space and are thus implicitly coupled through shared parameters, $r_d(a|q)$ does not directly depend on the parameters of the continuous branch. In practice, when generating valid continuous 3D poses, the discrete answers $q$ are highly templated. Thus, $r_d(a|q)$ can be treated as a constant with respect to the optimization of the continuous component. Therefore, the continuous policy gradient is proportional to:
\begin{equation}
    \nabla_{\theta} J_c \propto  \nabla_{\theta}\mathbb{E}_{q \sim \mathcal{D}, \left\{a_i, p_i\right\}_{i=1}^G \sim \pi_{\theta}(\cdot \mid q) }[r_c(q,a_i,p_i)\hat \Delta_i (q,a_i,p_i)].
\end{equation}

Based on the decomposed gradient structure, we apply PPO-style \citep{schulman2017proximal} clipping separately to the discrete and continuous components to stabilize training:
\begin{equation}
\begin{aligned}
\mathcal{J}_{\text{HyGRPO}} = 
& \mathbb{E}_{(q,a,p)\sim\mathcal{D}, \left\{a_i, p_i\right\}_{i=1}^G \sim \pi_{\theta}(\cdot \mid q)}
\Bigg[
\frac{1}{G} \sum_{i=1}^G \Big(
\min(r_d \hat{F}_i, \text{clip}(r_d, 1{-}\epsilon, 1{+}\epsilon)\hat{F}_i) \Big) \\
&+ \frac{1}{V} \sum_{i=1}^V \Big(\min(r_c \hat{\Delta}_i, \text{clip}(r_c, 1{-}\epsilon, 1{+}\epsilon)\hat{\Delta}_i) \Big)
- \beta D_{\text{KL}}(\pi_\theta \| \pi_{\text{ref}})
\Big) \Bigg].
\end{aligned}
\end{equation}

where $G$ is the total number of sampled candidates per input, and $V \leq G$ is the number of candidates with valid continuous outputs. This objective enables separate, stable, and reward-aligned optimization of discrete and continuous policy branches within a unified reinforcement learning framework.

\section{Experimental Details}
\label{appendix3}
The detailed hyperparameter settings for both Pose-specific MLLM pretraining and reinforcement fine-tuning are provided in Table~\ref{tab4}. In the pretraining stage, we focus on adapting the base LLaVA \citep{liu2023visual} model to 3D pose tasks, while the reinforcement fine-tuning stage further optimizes the policy behavior.

\begin{table}
\caption{Hyperparameter settings of pose-specific MLLM pretraining and reinforcement fine-tuning.}
  \centering
  \begin{tabular}{l|c|c}
    \toprule
    \textbf{Hyperparameters} & \multicolumn{1}{c|}{\textbf{Pretraining}}  & \multicolumn{1}{c}{\textbf{Reinforcement fine-tuning}} \\
    \midrule
    Batch Size & 80 & 16 \\
    Learning Rate & 3e-4 & 1e-6 \\
    Training Steps & 10000 & 1000 \\
    Optimizer & AdamW & AdamW \\
    Adam $\beta$ & (0.9, 0.95) & (0.9, 0.95) \\
    LR Schedule & Cosine & Cosine \\
    Computing Resources & NVIDIA A100 (40GB) & NVIDIA A800 (80GB)\\
    \bottomrule
  \end{tabular}
\label{tab4}
\end{table}

\section{Limitations}
\label{appendix4}
While our method represents a promising step toward reinforcement learning-based 3D human pose generation, it has several limitations. First, its effectiveness is inherently constrained by the quality of the reward functions. Designing reliable and semantically meaningful reward signals for pose generation remains a challenging problem, especially when capturing nuanced human preferences such as plausibility, naturalness, or contextual relevance. Inaccurate or incomplete reward feedback may misguide policy optimization, leading to suboptimal or unnatural poses.

Second, our framework relies on sampling multiple candidate responses per input to perform group-wise reward normalization. Although this design improves training stability in hybrid action spaces, it introduces non-negligible computational overhead, which may limit scalability when applied to larger models or datasets.

%%%%%%%%%%%%%%%%%%%%%%%%%%%%%%%%%%%%%%%%%%%%%%%%%%%%%%%%%%%%

\end{document}